\documentclass{article}
\usepackage{amsfonts,amsmath,amsthm,tikz,xcolor,mathtools,extarrows,graphicx,stackrel}
\usepackage[normalem]{ulem}
\useunder{\uline}{\ul}{}
\usepackage{hyperref}
\usepackage{colortbl}
\usepackage{hhline}
\usepackage{xcolor}
\usepackage{dsfont}
\usepackage{standalone}
\usepackage{comment}
\usepackage{wasysym} % for full and new moons
\usepackage{bbm} % for blackboard bold 1, k...
\usepackage{stmaryrd} % for Kh brackets
\usepackage{enumitem}
\usepackage{xargs}
\usepackage{color}
\usepackage[all, color, cmtip]{xy}
\colorlet{darkblue}{blue!70!black}
\colorlet{darkred}{red!70!black}
\colorlet{darkgreen}{green!70!black}
\colorlet{darkwhite}{white!65!black}
\colorlet{darkorange}{orange!70!black}

\colorlet{darkmagenta}{magenta!70!black}
\colorlet{pink}{green!20!magenta}

\colorlet{lightcyan}{cyan!15!white}
\colorlet{lightyellow}{yellow!30!white}
\colorlet{lightcyanb}{lightcyan!50!black}

\definecolor{purple}{rgb}{0.63, 0.36, 0.94} % official vertex color

\usetikzlibrary{decorations.markings,decorations.pathreplacing, decorations.pathmorphing, calligraphy,positioning,calc,arrows.meta,cd}

\usepackage{tikz}
\usetikzlibrary{shapes.geometric, arrows.meta, positioning, calc}

\definecolor{moduleblue}{RGB}{235, 245, 255}
\definecolor{operatorgreen}{RGB}{230, 250, 240}
\definecolor{featureorange}{RGB}{255, 245, 230}
\definecolor{darkblue}{RGB}{0, 50, 100}
\definecolor{darkgreen}{RGB}{0, 80, 40}
\definecolor{darkorange}{RGB}{120, 60, 0}
\numberwithin{equation}{section}
\theoremstyle{definition}

\usepackage{bbm}
\def\C{{\mathbb{C}}}

\def\R{{\mathbbm R}}

 \def\1{\mathbbm{1}}%

\renewcommand{\to}{\rightarrow}

\usepackage[preprint]{neurips_2026}

\usepackage[utf8]{inputenc} % allow utf-8 input
\usepackage[T1]{fontenc}    % use 8-bit T1 fonts
\usepackage{hyperref}       % hyperlinks
\usepackage{url}            % simple URL typesetting
\usepackage{booktabs}       % professional-quality tables
\usepackage{amsfonts}       % blackboard math symbols
\usepackage{nicefrac}       % compact symbols for 1/2, etc.
\usepackage{microtype}      % microtypography
\usepackage{xcolor}         % colors

\title{Analytic Torsion and Spectral Gap Capture Persistent-Laplacian Performance}

\author{%
  Jernej Grlj \\
  Department of Mathematics\\
  University of Southern California\\
  Los Angeles, CA 90089 \\
  \texttt{grlj@usc.edu} \\
  \And
  Aaron D. Lauda\\
  Department of Mathematics\\
  Department of Physics\\
  University of Southern California\\
  Los Angeles, CA 90089 \\
  \texttt{lauda@usc.edu} \\
  }

\begin{document}

\maketitle

\begin{abstract}
While persistent Laplacians (PL) offer a richer geometric representation of data than persistent homology, utilizing their full eigenspectrum for learning tasks is often hampered by high dimensionality and the ``varying length'' problem across different filtration scales. We propose a compact spectral representation that distills the persistent Laplacian into three mathematically grounded invariants: Betti numbers, the spectral gap, and analytic torsion. Across benchmark datasets including MNIST, QM-3D, and SKEMPI WT, we demonstrate that this reduced feature space captures the essential predictive signal of the full spectrum, and in some cases outperforms it, while significantly reducing computational overhead and preventing the noise introduced by higher-frequency eigenvalues. Our results suggest that these invariants provide a principled, fixed-length interface between spectral geometry and topological learning. 
\end{abstract}

\section{Introduction}
Persistent homology has become one of the central tools of topological data analysis, because it tracks the birth and death of homological features across scale. Its strength is also its limitation: by design, it records topology up to homological persistence, but not the finer geometric information that can continue to evolve even when the barcode does not. Once a cycle is born, for example, cycle size, heterogeneity, and internal organization may change substantially while its persistent homology remains unchanged. In this regime, Laplacian-based methods become a natural extension. They retain topological information within their harmonic sector, while their non-harmonic spectra record additional geometric and combinatorial structure invisible to barcodes alone \cite{EdelsbrunnerLetscherZomorodian2002, Mmoli2022, WeiWei2025}.

% This quantity is the higher-order analogue of algebraic connectivity in spectral graph
% theory. For graphs, the smallest nonzero Laplacian eigenvalue measures global
% connectivity and transport~\cite{Fiedler1973}; in our setting, the spectral gap plays the same conceptual
% role for the persistent \(q\)-Laplacian~\cite{SchaubEtAl2020}. It marks the first nontrivial
% frequency above the harmonic sector and therefore the slowest nontrivial diffusion scale
% visible to the operator. Small gap values indicate that a non-harmonic mode can be
% excited at low energy, which often reflects weak connectivity, near-degeneracy, or the
% presence of large-scale structure. Larger gap values indicate that the first nontrivial mode
% is more strongly separated from the kernel.

A useful way to frame this is through the classical spectral question posed by Kac: ``Can one hear the shape of a drum?'' \cite{Kac1966}. While the spectrum does not uniquely determine shape~\cite{GordonWebbWolpert1992}, it encodes a remarkable breadth of geometry and dynamics.  In Riemannian geometry, graph theory, and manifold learning, Laplace spectra govern oscillation, heat flow, diffusion, and multiscale structure. On graphs, the first nonzero eigenvalue already detects global connectivity, transport, and underlies spectral partitioning \cite{Fiedler1973}. Physically, it marks the first nontrivial frequency above the harmonic sector and therefore the slowest nontrivial diffusion scale visible to the operator. Small gap values indicate that a non-harmonic mode can be excited at low energy, which often reflects weak connectivity, near-degeneracy, or the presence of large-scale structure \cite{Fiedler1973,chung1997spectral}.   More broadly, Laplacian eigenvalues and eigenvectors drive diffusion maps, Laplacian eigenmaps, and random-walk-based representations of data \cite{BelkinNiyogi2003,CoifmanEtAl2005,SchaubEtAl2020}.  
%Low frequencies capture large-scale transport and coarse organization; higher frequencies probe increasingly fine oscillatory structure. 
From an algorithmic perspective, this means that the upper part of the spectrum is not decorative. It often carries discriminative information about local geometry and multiscale organization.

% This perspective extends naturally to simplicial complexes through Eckmann's combinatorial Laplacians \cite{Eckmann1945}. Persistent Laplacians (PL) transport this idea to filtrations $K \hookrightarrow L$, combining persistence with higher-order spectral analysis \cite{Mmoli2022}. While PLs have shown significant promise in biomolecular prediction and graph learning \cite{pmlr-v202-davies23c, WangNguyenWei2020}, their adoption faces a significant computational bottleneck.

This perspective extends naturally from graphs to simplicial complexes. Eckmann's combinatorial Laplacians generalize the graph Laplacian to operators acting on higher-dimensional simplices \cite{Eckmann1945}. In this setting, the kernel of the $q$-Laplacian recovers $q$-dimensional homology, while the nonzero eigenvalues capture additional structure beyond Betti numbers. Persistent Laplacians transport this idea to pairs of simplicial complexes $K \hookrightarrow L$, thereby combining persistence with higher-order spectral analysis \cite{Mmoli2022}. Their harmonic spectra recover persistent homology, but their non-harmonic spectra retain geometric information discarded by purely homological summaries \cite{WeiWei2025,WangNguyenWei2020}. This makes persistent Laplacians a natural candidate for topological learning pipelines that seek both multiscale topology and a quantitative description of how shape evolves across the filtration.  However, while these methods have shown significant promise in   biomolecular prediction and graph learning \cite{pmlr-v202-davies23c, WangNguyenWei2020}, their adoption faces a significant computational bottleneck.

As noted in recent evaluations \cite{pmlr-v202-davies23c, Xu2025}, packing spectral data into feature vectors is inherently difficult because the size of the spectra varies across different homological dimensions and filtration parameters. Previous attempts to resolve this typically rely on truncating the spectrum to a fixed number of eigenvalues or utilizing basic statistical summaries (e.g., mean or variance). However, naive truncation is often counterproductive: as the number of eigenvalues increases, the precision of the model often decreases \cite{pmlr-v202-davies23c}. This suggests that the higher part of the spectrum, while rich in information, often presents as an undifferentiated and noisy vector that hinders standard learning architectures.

Introduced by Ray and Singer as an analytic counterpart of Reidemeister torsion, analytic torsion is built from an alternating product of Laplacian pseudo-determinants and therefore depends on the full nonzero spectrum rather than only the harmonic sector \cite{RaySinger1971}. The subsequent work of Cheeger and M\"uller established the fundamental equivalence between analytic torsion and Reidemeister torsion in the classical setting \cite{Cheeger1979,Muller1978}. In particular, torsion is not an ad hoc spectral statistic: it is a mathematically principled bridge between topology and analysis, designed precisely to capture information beyond Betti numbers.

While analytic torsion has deep roots in topology and spectral geometry, it has not become a standard feature in topological data analysis. We argue that persistent Laplacians provide a natural setting in which to import this invariant: the harmonic sector recovers persistent homological information, and the spectral gap captures only the first positive eigenvalue. Meanwhile, the analytic torsion summarizes the contribution of \emph{all} nonzero eigenvalues. This makes torsion a mathematically grounded and computationally attractive feature for learning tasks built on persistent spectral data.

\subsection{Mathematical Feature Engineering.}\label{sec:math feature eng}
% \JG{removed intro paragraph}
The transition from raw spectral data to predictive performance is not merely a matter of dimensionality, but one of functional complexity. Extracting global geometric invariants from this spectrum, most notably the analytic torsion we study here, poses a significant challenge for generic machine learning models. Analytic torsion is a sophisticated, non-linear function of the entire non-harmonic spectrum, requiring the computation of alternating products of pseudodeterminants across varying homological degrees. For a model to ``discover" this invariant from raw eigenvalues, it would have to implicitly learn complex algebraic operations that are numerically sensitive and global in nature.
 
 By providing analytic torsion as a direct feature, we perform a form of mathematical feature engineering. We supply the model with high-level geometric primitives that would otherwise remain opaque within a high-dimensional, varying-length feature vector. Analytic torsion compresses the higher-frequency information into a single, principled quantity, allowing the model to bypass the noise and dimensionality issues inherent in raw spectral data.

For standard machine learning architectures, such as the Random Forests and Gradient Boosting Regressors utilized in this study, deriving this signal from a raw list of eigenvalues requires overcoming several significant functional hurdles:
\begin{itemize}[leftmargin=*]
    \item Isolating the harmonic sector: The model must correctly identify the varying number of zero eigenvalues (the Betti numbers) to determine the starting index of the non-harmonic spectrum.
    \item Computing multiplicative relationships: These models are inherently additive or based on axis-aligned splits; consequently, they struggle to approximate high-order multiplicative relationships, such as the product of all positive eigenvalues, without an explicit logarithmic transformation.
    \item Synthesizing across dimensions: The model would need to learn to integrate spectral data across multiple homological degrees $q$ while correctly applying the alternating sign convention and exponentiation required by the torsion formula.
\end{itemize}
Supplying these primitives directly performs algorithmic offloading for the model. This bypasses the learning bottleneck, preserves critical structural information, and provides a universal, fixed-length interface. This approach also eliminates the guesswork and manual tuning typically required by truncation-based methods to identify thresholds of diminishing returns

 \subsection{Our Contributions.}
 
 % \JG{removed intro sentence}The main contribution of this paper is threefold:

\paragraph{Resolution of the Spectral Packing Problem} We introduce a fixed-dimensional representation that resolves the inherent difficulty of utilizing varying-length spectral vectors across different homological dimensions and filtration parameters. This allows for a consistent input format for standard machine learning architectures without the need for heuristic truncation or basic statistical pooling.

\paragraph{A Principled Topological-Spectral Partitioning:} We propose a novel feature set grounded in the structural decomposition of the persistent Laplacian. By utilizing the harmonic sector (Betti numbers) to capture persistent topology and spectral gap and analytic torsion to summarize the non-harmonic spectrum, we provide a mathematically rigorous alternative to naive eigenvalue vectorization.

\paragraph{Cross-Domain Empirical Robustness:}
We demonstrate the effectiveness of our topological-spectral representation across a widely varying suite of benchmarks, ranging from quantum mechanical properties of small molecules to large-scale protein engineering tasks and manifold learning in image datasets. Our results show that this mathematical feature reduction remains effective regardless of the underlying data dimensionality or domain-specific noise, and effectively circumvents the breaking point observed in previous studies, where increasing the number of raw eigenvalues leads to diminishing returns or decreased model precision \cite{pmlr-v202-davies23c}. 
% \JG{removed sentence}We show that mathematical feature engineering, providing the model with high-level geometric primitives, preserves critical signal while significantly reducing dimensionality and noise.

% Ultimately, our work supports a simple thesis: the success of spectral methods in TDA rests on the fact that nonzero eigenvalues encode meaningful structure. By using analytic torsion to bridge the gap between topology and analysis, we offer a representation that is lower-dimensional, more interpretable, and computationally robust.
% \JG{maybe remove this paragraph?}
More broadly, our results support a simple thesis: the success of spectral methods in geometry, graph theory, and data analysis has always rested on the fact that nonzero eigenvalues encode meaningful structure. By utilizing the spectral gap to capture the leading nontrivial diffusion scale and analytic torsion to summarize the cumulative higher-frequency content, we offer a compact spectral interface between topology, geometry, and learning. This tripartite representation, comprising Betti numbers, the spectral gap, and analytic torsion, provides a principled, fixed-length summary that preserves the predictive power of the full persistent Laplacian while significantly reducing the noise and computational overhead associated with high-dimensional spectral data. In doing so, we establish a mathematically transparent framework for characterizing how shape and connectivity evolve across a filtration, effectively bridging the gap between classical spectral geometry and modern topological data analysis.

 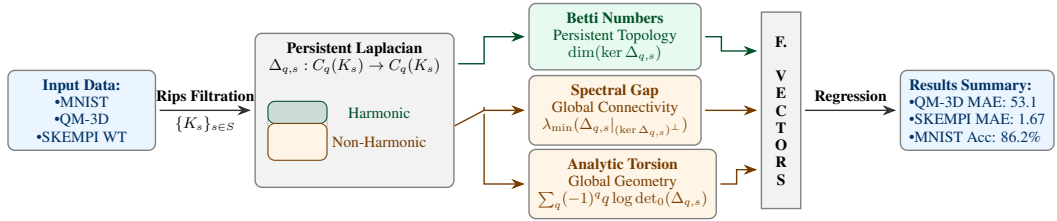
\begin{figure*}[t]
\centering
% This line forces the entire TikZ diagram to fit within the page margins
\resizebox{\textwidth}{!}{
    \begin{tikzpicture}[
        node distance=1.5cm and 1.2cm,
        base/.style={rectangle, rounded corners, draw=black!50, minimum width=3cm, minimum height=1cm, align=center, font=\small, inner sep=5pt},
        datanode/.style={base, fill=moduleblue, text=darkblue, thick},
        opnode/.style={base, fill=operatorgreen,  thick, minimum height=1.5cm},
        featnode/.style={base, fill=featureorange, text=darkorange, thick, minimum width=3.5cm},
        vecnode/.style={rectangle, draw=black!50, fill=black!5, thick, minimum width=0.8cm, minimum height=4cm, align=center, font=\small\bfseries, inner sep=2pt},
        mlnode/.style={diamond, draw=black!50, fill=black!5, thick, aspect=1.5, align=center, font=\small, inner sep=0pt},
        arrow/.style={-{Stealth[scale=1.2]}, thick, shorten >=2pt, shorten <=2pt, draw=black!80}
    ]
    \node (raw) [datanode] {\textbf{Input Data:} \\ 
    \textbullet MNIST \\
    \textbullet QM-3D \\ 
    \textbullet SKEMPI WT};
    % \node (mod1_label) [above=0.2cm of raw, font=\small\bfseries] {Module 1};
    % \node (filt) [datanode, right=of raw] {\textbf{Simplicial Filtration} \\ $\{K_s\}_{s \in S}$};
    % \node (filt_param) [below=0.1cm of filt, font=\footnotesize\itshape, text=darkblue] {Parameter $s$};
    \node (laplacian) [opnode,draw=black!50, fill=black!5, right=2cm of raw] {
        \textbf{Persistent Laplacian} \\ 
        $\Delta_{q,s}: C_q(K_s) \to C_q(K_s)$ \\
        \mbox{
        \begin{tikzpicture}[scale=0.3]
            \draw[draw=black!30, fill=white] (0,0) rectangle (4,4);
            \draw[fill=darkgreen!20, draw=darkgreen] (0,2.5) rectangle (4,4); 
            \node[text=darkgreen] at (2, 3.25) {\quad Harmonic};
            \draw[fill=featureorange, draw=darkorange] (0,0) rectangle (4,2.5); 
            \node[text=darkorange] at (2, 1.25) { \quad Non-Harmonic};
        \end{tikzpicture}}
    };
    % \node (mod2_label) [above=0.2cm of laplacian, font=\small\bfseries] {Module 2};
    \coordinate[right=0.6cm of laplacian.east] (split_coord);
   % === MODULE 3: FEATURE EXTRACTION (Tripartite) ===
% Increased vertical offsets (1.5cm) to prevent node stacking
\node (gap) [featnode, right=1.5cm of laplacian.east, anchor=west] {
    \textbf{Spectral Gap} \\ 
    Global Connectivity \\ 
    $\lambda_{\min}(\Delta_{q,s}|_{(\ker \Delta_{q,s})^\perp})$
};
\node (betti) [featnode, above=1.5cm of gap.west, anchor=west, fill=operatorgreen, text=darkgreen,] {
    \textbf{Betti Numbers}\\ 
    Persistent Topology \\ 
    $\dim(\ker \Delta_{q,s})$
};
\node (torsion) [featnode, below=1.5cm of gap.west, anchor=west] {
    \textbf{Analytic Torsion} \\ 
    Global Geometry \\ 
    $\sum_q (-1)^q q \log \det_0(\Delta_{q,s})$
};
% \node (mod3_label) [above=0.2cm of betti, font=\small\bfseries] {Module 3};
% Background fit adjusted for new spacing
% \node [draw=darkorange!30, dashed, fill=featureorange!20, rounded corners] {};
   \node (vector) [vecnode, right=1.2cm of gap.east, anchor=west] {F.\\ \\V\\E\\C\\T\\O\\R\\S};
    % \node (mod4_label) [above=0.2cm of vector, font=\small\bfseries] {Module 4};
    % \node (fixed_len) [below=0.1cm of vector, font=\footnotesize\bfseries, text=black!70, align=center] {Fixed\\Length};
    % \node (ml) [mlnode, right=1.2cm of vector] {\textbf{Random}\\\textbf{Forest}};
    % \node (mod5_label) [above=0.2cm of ml, font=\small\bfseries] {Module 5};
    \node (results) [datanode, right=2cm of vector, align=left] {
        \textbf{Results Summary:} \\
        \textbullet QM-3D MAE: 53.1 \\
        \textbullet SKEMPI MAE: 1.67 \\
        \textbullet MNIST Acc: 86.2\%
    };
    \draw [arrow] (raw) -- (laplacian)  node [midway, above, font=\small\bfseries] {Rips Filtration} node [midway, below, font=\small\bfseries] {$\{K_s\}_{s \in S}$};
    % \draw [arrow] (filt) -- (laplacian);
    \draw [arrow, darkgreen] ($(laplacian.north east)!0.2!(laplacian.south east)$) -- ++(0.6,0) |- (betti.west);
    \draw [thick, darkorange] ($(laplacian.north east)!0.6!(laplacian.south east)$) -- (split_coord);
    \draw [arrow, darkorange] (split_coord) |- (gap.west);
    \draw [arrow, darkorange] (split_coord) |- (torsion.west);
    \draw [arrow, darkgreen] (betti.east) -- ++(0.4,0) |- ($(vector.north west)!0.2!(vector.south west)$);
    \draw [arrow, darkorange] (gap.east) -- (vector.west);
    \draw [arrow, darkorange] (torsion.east) -- ++(0.4,0) |- ($(vector.north west)!0.8!(vector.south west)$);
    % \draw [arrow, ->, dashdotted, darkblue!70] (filt_param.south) -- ++(0,-0.8) -- ++(9.5,0) |- (vector.south);
    \draw [arrow] (vector.east) -- (results) node [midway, above, font=\small\bfseries] {Regression};
    % \draw [arrow] (ml.east) -- (results);
    % \node[anchor=south west, font=\footnotesize, text=black!60] at ($(raw.north west)+(0,1.2)$) {
    %     \textbf{Color Key:} \hspace{1em} 
    %     {\large\color{moduleblue}\textbullet} Data Handling \hspace{1em}
    %     {\large\color{operatorgreen}\textbullet} Spectral Operator \hspace{1em}
    %     {\large\color{featureorange}\textbullet} Tripartite Contribution
    % };
    \end{tikzpicture}
} % End of resizebox
\caption{Conceptual pipeline for mathematical feature engineering using persistent Laplacians. The multi-scale persistent Laplacian $\Delta_{q,s}$, formed from Rips complexes, is decomposed into its harmonic and non-harmonic components. These contribute to the tripartite feature vectors consisting of Betti numbers, spectral gap, and analytic torsion. The results are obtained using appropriate regression methods.}
\label{fig:pipeline_concept}
\end{figure*}

\subsection*{Organization of the paper}
% \JG{this can be removed in my opinion} 
The remainder of this paper is organized as follows: Section~\ref{sec:related} reviews related work in topological and spectral learning; Section~\ref{sec:prelim} provides the necessary mathematical preliminaries on persistent Laplacians and torsion; Section~\ref{sec:exp} details our feature construction and computational considerations; and Section~\ref{sec:benchmark} presents our experimental results, ablation studies, and training times across various benchmarks.
 
% ############################################
% ############################################
\subsection{Related Work} \label{sec:related}
% ############################################
% ############################################

\paragraph{Vectorization of Topological Summaries:}
While persistent homology provides a robust multiscale summary of data \cite{Carlsson2009}, its raw outputs of persistence diagrams and barcodes are not natively compatible with standard machine learning kernels \cite{pmlr-v202-davies23c}. This limitation has motivated several vectorization frameworks, most notably Persistence Landscapes~\cite{bubenik2015statistical},  Persistence Images ~\cite{adams2017persistence}, and Persistence Silhouettes~\cite{chazal2015stochastic}. These methods typically focus on functional or density-based summaries of birth-death pairs. While successful in many applications, these representations remain strictly homological; they inherit the limitation of ignoring the finer geometric and combinatorial evolution that occurs even when the barcode remains static. Our work with spectral invariants seeks to fill this gap by probing the geometry of the filtration that these density-based summaries omit.

\paragraph{Spectral Methods and Higher-Order Structure}
Our approach sits within a well-established tradition of using Laplacian spectra for representation learning in manifold learning \cite{BelkinNiyogi2003, CoifmanEtAl2005},  out-of-distribution robustness diagnostics \cite{zia2026representationgeometrydiagnosticoutofdistribution}, and graph theory \cite{Fiedler1973}. Foundational methods such as Laplacian eigenmaps and diffusion maps leverage low-frequency eigenvalues to detect global connectivity and coarse organization. These concepts extend from graphs to higher-order networks through Eckmann’s combinatorial Laplacians \cite{Eckmann1945}, where nonzero eigenvalues capture structural information and higher-order diffusion that Betti numbers alone cannot resolve \cite{SchaubEtAl2020}. By extracting global invariants from these operators, we leverage this broader spectral-learning tradition to characterize multiscale organization in simplicial complexes.

\paragraph{Persistent Laplacians and Spectral Representations}
This paper builds on the persistent Laplacian (PL) framework introduced by M\'{e}moli et al. \cite{Mmoli2022}, which tracks the evolution of the non-harmonic spectrum across a filtration.  Extensions of this work include the Persistent sheaf Laplacians \cite{jones2026petlspersistenttopologicallaplacian, wang2026multidimensionalpersistentsheaflaplacians, Wei2025}, persistent path Laplacians \cite{Wang2023} and persistent hyperdigraph Laplacians \cite{Chen2023} for instance.  Persistent Laplacians have already shown significant promise in practical applications, such as the projection of COVID-19 variants \cite{Chen2022} and protein-protein interaction prediction \cite{Xu2025, WangNguyenWei2020}. However, utilizing the full spectrum for learning introduces the ``spectral packing'' problem: the size of eigenspectra varies significantly across homological dimensions and filtration parameters \cite{pmlr-v202-davies23c}. Previous attempts to resolve this have relied on truncating the spectrum to a fixed number of eigenvalues \cite{pmlr-v202-davies23c} or utilizing basic statistical descriptors like mean and standard deviation \cite{Xu2025}. In a different direction \cite{jung2025persistentlaplaciandiagrams} studies the persistent Laplacian image as the vectorization of spectral information, generalizing \cite{adams2017persistence}.  Our results contribute to this effort by demonstrating that the spectral content can be compressed into principled invariants, Betti numbers, gaps, and torsion, without the noise or diminishing returns associated with naive truncation.

\paragraph{Torsion-Based Summaries}
Analytic torsion has deep roots in spectral geometry as a bridge between topology and analysis, originally developed by Ray and Singer \cite{RaySinger1971} and further established by Cheeger and M\"{u}ller \cite{Cheeger1979, Muller1978}. While historically used for the classification of manifolds \cite{Reidemeister1935}, its use as a feature in modern learning pipelines remains an emerging frontier. 
Recent work has begun to integrate torsion into graph neural networks (GNNs) as a local weighting device within message-passing protocols \cite{shen2025torsion, li2025geometry}. Our approach represents a distinct shift in methodology: rather than utilizing torsion for local architectural weighting, we treat it as a \textit{global invariant} extracted directly from the persistent Laplacian. This allows us to import a classical topological invariant into a persistent spectral learning framework in a way that is both mathematically natural and algorithmically useful.

\section{Preliminaries on Persistent Laplacians and Spectral Summaries} \label{sec:prelim}
% ############################################
% ############################################

We present some minimal mathematical background needed to understand the problem at hand. For more details on persistent Laplacians, we refer the reader to \cite{Mmoli2022}, and to \cite{Carlsson2009} for the original work on persistent homology. 

% -------------------------------------------
\subsection{Simplicial complexes, chain groups, and boundary operators}
% -------------------------------------------

A (finite, abstract) simplicial complex $K$ on a vertex set $V$ is a collection of finite subsets of $V$
that is closed under taking subsets: if $\sigma \in K$ and $\tau \subseteq \sigma$, then $\tau \in K$.
If $\sigma$ has $q+1$ vertices, then $\sigma$ is called a \emph{$q$-simplex}, and we write $K_q$ for the
set of all $q$-simplices in $K$. Thus, $K_0$ is the set of vertices, $K_1$ the set of edges, $K_2$ the
set of triangles, and so on.

To define Laplacians, one passes from simplices to oriented chain groups. For each $q \geq 0$, let
$C_q(K)$ denote the vector space of formal linear combinations of oriented $q$-simplices in $K$ with
coefficients in $\mathbb{R}$. Equivalently, after choosing an orientation for each simplex in $K_q$,
the group $C_q(K)$ may be identified with $\mathbb{R}^{|K_q|}$. We write a typical basis element as
$
[v_0,v_1,\dots,v_q],
$
where the ordering specifies the orientation, and reversing orientation changes sign in the usual way.

The boundary operator
$
\partial_q : C_q(K) \to C_{q-1}(K)
$
is defined on an oriented simplex by
\[
\partial_q [v_0,\dots,v_q]
=
\sum_{i=0}^q (-1)^i [v_0,\dots,\widehat{v_i},\dots,v_q],
\]
where $\widehat{v_i}$ indicates omission of the vertex $v_i$, and the map is extended linearly.
The fundamental identity
$
\partial_{q-1}\partial_q = 0
$
holds for all $q$, so the sequence
\[
\cdots \xrightarrow{\partial_{q+1}} C_q(K) \xrightarrow{\partial_q} C_{q-1}(K)
\xrightarrow{\partial_{q-1}} \cdots \xrightarrow{\partial_1} C_0(K) \to 0
\]
forms a chain complex.
The $q$-th homology group is then
\[
H_q(K) = \ker(\partial_q)/\operatorname{im}(\partial_{q+1}),
\]
which records $q$-dimensional cycles modulo boundaries. In the present paper, this topological information
will appear as the harmonic part of the Laplacian, while the nonzero spectrum will encode additional
geometric and combinatorial structure.

% \JG{remove this paragraph}For computation, it is convenient to represent $\partial_q$ as a matrix $B_q$ once an ordering and
% orientation of simplices have been fixed. The entries of $B_q$ lie in $\{0,\pm 1\}$, indicating whether
% a $(q-1)$-simplex appears as a face of a $q$-simplex with matching or opposite orientation. These boundary
% matrices are the basic algebraic input from which both combinatorial and persistent Laplacians are built.

When working with a filtration
$
K^{(0)} \subseteq K^{(1)} \subseteq \cdots \subseteq K^{(T)},
$
or more generally, with an inclusion of simplicial complexes $K \hookrightarrow L$, the corresponding chain
groups and boundary operators are related by restriction to the appropriate simplex sets. This viewpoint
will allow us to define persistent Laplacians in the next subsection.

% -------------------------------------------
\subsection{Combinatorial Laplacians and Hodge decomposition}
% -------------------------------------------

Given a simplicial complex $K$, we equip each chain group $C_q(K)$ with the standard Euclidean inner
product in the oriented simplex basis. With respect to this inner product, the adjoint of the boundary
operator
$
\partial_q : C_q(K) \to C_{q-1}(K)
$
is the map
$
\partial_q^* : C_{q-1}(K) \to C_q(K).
$
In matrix form, if $B_q$ denotes the boundary matrix, then $\partial_q^*$ is represented by $B_q^\top$.

The \emph{combinatorial $q$-Laplacian} is defined by
\[
\Delta_q
=
\partial_{q+1}\partial_{q+1}^* + \partial_q^*\partial_q.
\]
% \JG{removed paragraphs}
% Equivalently, one may write
% \[
% \Delta_q = \Delta_q^{\mathrm{up}} + \Delta_q^{\mathrm{down}},
% \qquad
% \Delta_q^{\mathrm{up}} := \partial_{q+1}\partial_{q+1}^*,
% \qquad
% \Delta_q^{\mathrm{down}} := \partial_q^*\partial_q,
% \]
% where the up-Laplacian couples $q$-simplices through shared $(q+1)$-cofaces, while the down-Laplacian
% couples them through shared $(q-1)$-faces. In the oriented simplex basis, this becomes
% \[
% L_q = B_{q+1}B_{q+1}^\top + B_q^\top B_q.
% \]

% The operator $\Delta_q$ is symmetric and positive semidefinite. Its kernel consists of those $q$-chains
% that are both cycle-like and cocycle-like:
% \[
% \ker(\Delta_q) = \ker(\partial_q)\cap\ker(\partial_{q+1}^*).
% \]
% These are the \emph{harmonic $q$-chains}. A basic consequence of discrete Hodge theory is that harmonic
% chains represent homology classes, and indeed there is a canonical isomorphism
The elements of $\ker(\Delta_q)$ are the \emph{harmonic $q$-chains}. A basic consequence of discrete Hodge theory is that harmonic
chains represent homology classes, and indeed, there is a canonical isomorphism
\[
\ker(\Delta_q) \cong H_q(K).
\]
Thus, the zero eigenspace of the combinatorial Laplacian recovers the topological information encoded by
$q$-dimensional homology.

% \JG{remove this paragraph}More generally, one has the Hodge decomposition
% \[
% C_q(K)
% =
% \operatorname{im}(\partial_{q+1})
% \oplus
% \ker(\Delta_q)
% \oplus
% \operatorname{im}(\partial_q^*).
% \]
% This decomposition separates $q$-chains into boundaries, harmonic representatives, and coexact components.
% For the purposes of this paper, the important point is that topology lives in the harmonic sector,
% that is, in the kernel of the Laplacian, while the nonzero spectrum captures additional combinatorial and
% geometric structure not visible from homology alone.

% This distinction is central to Laplacian-based learning. 

The harmonic component detects persistent
topological features, while
% but
% \JG{changed from but to while} 
the positive eigenvalues measure how $q$-simplices interact through the complex.
% \JG{removed, repeated in intro}These eigenvalues govern higher-order diffusion and oscillation, in direct analogy with spectral methods on
% graphs and manifolds~\cite{BelkinNiyogi2003,SchaubEtAl2020}. In particular, lower nonzero eigenvalues reflect coarse global organization, while
% larger eigenvalues probe finer-scale structure. Persistent Laplacians build on this picture by retaining the
% homological information in the kernel while tracking how the non-harmonic spectrum evolves across a
% filtration.

% -------------------------------------------
\subsection{Persistent Laplacians}
% -------------------------------------------
Persistent homology tracks which homological features of a smaller complex remain visible
inside a larger one. Let \(X \subseteq Y\) be simplicial complexes, for example two scales
\(X = K^{(s)}\) and \(Y = K^{(t)}\) in a filtration with \(s \le t\). Intuitively, the
\(q\)-dimensional persistent homology of the (filtration) pair \((X,Y)\) consists of those \(q\)-cycles
in \(X\) that remain nontrivial when viewed inside \(Y\). Formally,
\[
H_q^{X,Y}
:=
\ker(\partial_q^X)\Big/\left(\operatorname{im}(\partial_{q+1}^Y)\cap C_q(X)\right).
\]

To define the corresponding persistent Laplacian, one restricts attention to those
\((q+1)\)-chains in \(Y\) whose boundary lies in \(C_q(X)\). Specifically, define
\[
C_{q+1}^{X,Y}
:=
\left\{ c \in C_{q+1}(Y) \;:\; \partial_{q+1}^Y c \in C_q(X) \right\},
\]
and let
$
\partial_{q+1}^{X,Y} : C_{q+1}^{X,Y} \to C_q(X)
$
denote the restriction of \(\partial_{q+1}^Y\) to this subspace. Since \(C_{q+1}^{X,Y}\)
inherits an inner product from \(C_{q+1}(Y)\), the adjoint
\[
(\partial_{q+1}^{X,Y})^* : C_q(X) \to C_{q+1}^{X,Y}
\]
is well defined. The \(q\)-th persistent Laplacian is then
\[
\Delta_q^{X,Y}
=
\partial_{q+1}^{X,Y}(\partial_{q+1}^{X,Y})^*
+
(\partial_q^X)^*\partial_q^X.
\]
% It is natural to separate this into an up and down part: \JG{remove?}
% \[
% \Delta_q^{X,Y}
% =
% \Delta_{q,+}^{X,Y}+\Delta_{q,-}^{X,Y},
% \quad
% \Delta_{q,+}^{X,Y}
% :=
% \partial_{q+1}^{X,Y}(\partial_{q+1}^{X,Y})^*,
% \quad
% \Delta_{q,-}^{X,Y}
% :=
% (\partial_q^X)^*\partial_q^X.
% \]

The key structural fact is the persistent Hodge theorem:
$
\ker(\Delta_q^{X,Y}) \cong H_q^{X,Y}.
$
Thus, the harmonic sector of the persistent Laplacian recovers exactly the persistent
homological information of the pair. In particular, the multiplicity of the zero eigenvalue
is the persistent Betti number. This can be understood as the definition of persistent homology. When \(X=Y\), this reduces to the ordinary combinatorial
Laplacian and ordinary homology.

% \JG{removed, repeated in intro}The advantage of persistent Laplacians is that they do more than recover persistence.
% The zero eigenspace still encodes topology, but the positive eigenvalues retain additional
% information about how the complex is organized across scales. In this sense, persistent
% Laplacians refine persistent homology: the harmonic spectrum recovers the barcode-level
% information, while the non-harmonic spectrum records geometric and combinatorial structure
% that evolves along the filtration.

% This makes persistent Laplacians especially natural for topological learning pipelines.
% Persistent homology provides a robust multiscale topological summary, but it discards
% spectral information once one passes to homology classes alone. Persistent Laplacians retain
% that topological core while adding a non-harmonic spectral channel that can be used for
% representation learning. In particular, they provide a principled setting in which one can
% study which parts of the spectrum matter most algorithmically, and whether compact
% summaries of that spectrum can recover much of the performance of full spectral features
% \cite{Mmoli2022,WeiWei2025}.

% -------------------------------------------
\subsection{Spectral gap}
% -------------------------------------------
Denote the eigenvalues $
0 \leq \lambda_{q,1}^{X,Y} \le \lambda_{q,2}^{X,Y} \le \cdots \le \lambda_{q,n_q}^{X,Y}
$ of the persistent \(q\)-Laplacian \(\Delta_q^{X,Y}\) listed in
nondecreasing order and repeated according to multiplicity. Since \(\Delta_q^{X,Y}\) is
symmetric and positive semidefinite, all eigenvalues are real and nonnegative.
Let
\[
m = \dim\ker(\Delta_q^{X,Y}) = \beta_q^{X,Y}
\]
denote the persistent Betti number. Then
$
\lambda_{q,1}^{X,Y}=\cdots=\lambda_{q,m}^{X,Y}=0
$
when \(m >0\), while \(\lambda_{q,1}^{X,Y}>0\) when \(m=0\). 
%\JG{removed line break}
We define the \emph{spectral gap} of \(\Delta_q^{X,Y}\) to be
\[
\gamma_q^{X,Y}
:=
\min\{\lambda>0:\lambda\in\mathrm{spec}(\Delta_q^{X,Y})\},
\]
provided that \(\Delta_q^{X,Y}\) has at least one positive eigenvalue. Equivalently, if
\(m < n_q\), then
$
\gamma_q^{X,Y}=\lambda_{q,m+1}^{X,Y}.
$
Thus the spectral gap is the smallest nonzero eigenvalue of the persistent Laplacian.

\subsection{Analytic torsion} \label{subsec:analytic}
% -------------------------------------------

For more details, we refer the reader to \cite{mnev2014lecturenotestorsions}. The spectral gap isolates only the first positive eigenvalue of the persistent Laplacian.
To summarize the rest of the non-harmonic spectrum, we use a determinant-type spectral
statistic derived from analytic torsion.

% \JG{remove this paragraph}In the smooth setting of Riemannian geometry, Ray--Singer analytic torsion is defined from zeta-regularized
% determinants of Hodge Laplacians. In the light of Cheeger-M\"uller theorem \cite{Cheeger1979,Muller1978}, analytic torsion relates to the classification of lens spaces \cite{Reidemeister1935}. In the finite-dimensional simplicial setting considered
% here, the corresponding discrete quantity is expressed in terms of pseudodeterminants,
% that is, products of nonzero eigenvalues. This provides a natural way to compress the
% entire positive spectrum of a Laplacian into a single scalar.

Let
$
\lambda_{q,1}^{X,Y} \le \lambda_{q,2}^{X,Y} \le \cdots \le \lambda_{q,n_q}^{X,Y}
$
be the eigenvalues of the persistent \(q\)-Laplacian \(\Delta_q^{X,Y}\), and let
\[
m_q=\dim\ker(\Delta_q^{X,Y})
\]
be the persistent Betti number in degree \(q\). 
% \JG{removed to log only}
% We define the pseudodeterminant of
% \(\Delta_q^{X,Y}\) by
% \[
% \det\nolimits_{0}(\Delta_q^{X,Y})
% :=
% \prod_{i=m_q+1}^{n_q} \lambda_{q,i}^{X,Y},
% \]
% with the convention that \(\det_{0}(\Delta_q^{X,Y})=1\) if there are no positive eigenvalues.
% Equivalently,
% \[
% \log \det\nolimits_{0}(\Delta_q^{X,Y})
% =
% \sum_{\lambda \in \mathrm{spec}(\Delta_q^{X,Y}),\, \lambda>0} \log \lambda.
% \]
% Thus, zero eigenvalues do not contribute to the pseudodeterminant; they are excluded
% rather than regularized or perturbed. This is the finite-dimensional analogue of taking
% the determinant over the non-harmonic sector only.
We define the (log-)pseudodeterminant of
\(\Delta_q^{X,Y}\) by
% \[
% \det\nolimits_{0}(\Delta_q^{X,Y})
% :=
% \prod_{i=m_q+1}^{n_q} \lambda_{q,i}^{X,Y},
% \]
% with the convention that \(\det_{0}(\Delta_q^{X,Y})=1\) if there are no positive eigenvalues.
% Equivalently,
\[
\log \det\nolimits_{0}(\Delta_q^{X,Y})
=
\sum_{\lambda \in \mathrm{spec}(\Delta_q^{X,Y}),\, \lambda>0} \log \lambda.
\]
Thus, zero eigenvalues do not contribute to the pseudodeterminant; they are excluded
rather than regularized or perturbed. This is the finite-dimensional analog of taking
the determinant over the non-harmonic sector only.

Using these pseudodeterminants, a discrete torsion statistic may be defined by the
alternating product
%\JG{removed non log torsion}
% \[
% T(X,Y)
% :=
% \prod_{q \ge 0} \det\nolimits_{0}(\Delta_q^{X,Y})^{\frac{(-1)^{q+1}q}{2}},
% \]
% or equivalently in logarithmic form,
\[
\log T(X,Y)
=
\frac{1}{2}\sum_{q \ge 0} (-1)^{q+1} q \,
\log \det\nolimits_{0}(\Delta_q^{X,Y}).
\]
This is the natural finite-dimensional torsion-style summary associated with the persistent
Laplacians of the pair \((X,Y)\). 
% \JG{removed torsion feature in one degree}
% In settings where only a single degree \(q\) is used in the
% feature pipeline, the relevant quantity reduces to the degree-\(q\) contribution
% \[
% \tau_q^{X,Y} := \log \det\nolimits_{0}(\Delta_q^{X,Y}),
% \]
% which we will refer to as the torsion feature in that degree.

For a (weighted) graph $G$, the definition of torsion reduces to the notion of graph determinant (note that we removed the factor of $q$)
% \JG{removed to log}
% \begin{align*}
%      T_G(X,Y)
% =
% % \frac{1}{n_0 - m_0 + 1} \,
% \det\nolimits_{0}(\Delta_0^{X,Y}).
% \end{align*}
\begin{align*}
    \log T_G(X,Y)
=
% \frac{1}{n_0 - m_0 + 1} \,
\log \det\nolimits_{0}(\Delta_0^{X,Y}).
\end{align*}
By Kirchhoff's theorem $ \frac{1}{n_0 - m_0 + 1} \log T_G(G,G)$ counts the number of spanning trees in $G$. In this zero-dimensional setting, $\Delta_0^{G,G}$ agrees with the standard definition of a graph Laplacian using the difference of degree and adjacency matrices. See \cite{Moore2011-rc} for details.

\section{Experimental Setup} \label{sec:exp}
% ############################################
% ############################################

Using the mathematics that we have introduced, we explain our feature extraction from Laplacians.
% \JG{added}
% For clarity, we provide dataset specific comments.

\subsection{Feature construction}

We construct our feature vectors as follows. For each filtration pair, we append the Betti numbers in all homological dimensions, the spectral gaps in all homological dimensions, and the analytic torsion (we use the logarithmic form for computational stability) to the corresponding feature vector. In practice, this gets modified slightly depending on the dataset, as explained in the next section.

\subsection{Benchmark datasets}

To demonstrate the universal nature of analytic torsion and spectral gap as primary features driving persistent Laplacians, we examine how the features provide advantage across a broad spectrum of benchmarking data sets. 
% \JG{removed sentence}In particular, we examine the existing work of \cite{Xu2025} for protein-protein interaction on SKEMPI WT dataset \cite{Moal2012}, and \cite{pmlr-v202-davies23c} for integer recognition on the MNIST dataset \cite{lecun1998mnist}, and molecule energy on the QM-3D dataset \cite{Rupp2012, Blum2009}.
The datasets are chosen to demonstrate the efficiency of our feature vectors across $ 1$, $2$, and $ 3$-dimensional data. We briefly describe the experimental setup for each dataset. The detailed description of how the persistent Laplacians are constructed from the relevant dataset can be found in the corresponding papers, whose code is openly available (see Section \ref{sec:code}).
\smallskip

\noindent \textbf{SKEMPI WT}(wild-type) dataset provides the baseline binding affinity and structural data for $343$ unmutated protein-protein complexes. It is a subset of the comprehensive SKEMPI v2 dataset (CC BY 4.0) \cite{Moal2012, 10.1093/bioinformatics/bty635}, which catalogs thermodynamic and kinetic data for thousands of mutated protein interactions. The data is used to predict the binding energy of the protein-protein interactions, values in our work are in kcal/mol. Importantly for our application, the raw data includes spatial positions and names of elements on the binding interface.

% \JG{remove parahaph}
% By splitting the elements at the binding interface into two sets, we construct a modified distance function where the atoms from the same group are considered infinitely far apart (for the purpose of coding, infinite distance corresponds to $-1$), see \cite[Section 2.1.2]{Xu2025} for details. Using this modified distance a weighted graph is constructed, where the vertices are the atoms and weights of edges the distances between them.

% This graph is then used to construct Rips and Alpha complexes \cite{Edelsbrunner1994, Vietoris1927} and extract features from those using standard approaches, see \cite{Xu2025}.
We construct Rips and Alpha complexes from the data \cite{Edelsbrunner1994, Vietoris1927} and extract features from those using standard approaches, see \cite{Xu2025}. We keep those features untouched. Additionally, we construct the persistent graph Laplacians and extract the zeroth Betti number, the spectral gap, and the graph torsion. We also extract the first Betti number using the two different interpretations of the Euler characteristic of the graph. Hence, the contribution to the feature vector from the Laplacian is four-dimensional. The original feature vector consisted of eight statistical quantities.
\smallskip

\noindent \textbf{MNIST} (GNU GPL v3.0) \cite{lecun1998mnist} is a benchmark computer vision dataset containing $70,000$ grayscale $28 \times 28$ images of handwritten digits. The data is used for training models on image (in this case number) recognition. We process the images using a single height filtration and by reducing the images from grayscale to black and white, as in \cite{pmlr-v202-davies23c}. Persistent Laplacians are formed from cubical complexes in dimension $0$ and $1$. This corresponds to the fact that persistent homology becomes trivial in dimensions equal to or higher than the dimensions of the object.

From the spectral data we extract $2\cdot 2 + 1$ dimensional feature vectors. These features are precisely those described at the beginning of this section. The original feature vectors had length $2 \cdot 10$ (for $10$ eigenvalues in $2$ dimensions).
\smallskip

\noindent \textbf{QM-3D} is a roughly $7,000$ element subset of the larger dataset MoleculeNet (MIT license) \cite{Rupp2012, Blum2009, Wu2018}. The subset that we use consists of position vectors of each element in a molecule. Other versions include additional information about the molecule or the atoms. The data is used for predicting the energy of the molecules measured in  kcal/mol. The data comes with an official $P$-matrix split for the training and testing data. Using the Rips complex \cite{Vietoris1927}, we obtain a simplicial complex from which we construct the persistent Laplacians. 

From the spectral data, we extract $3\cdot 2 + 1$ dimensional feature vectors. These features are precisely those described at the beginning of this section. The original feature vectors had length $3 \cdot 10$ (for $10$ eigenvalues in $3$ dimensions).

\subsection{Learning protocols}\label{sec:learning protocol}

% Specify the train/validation/test setup, model class, hyperparameter selection, number of runs, and evaluation metrics. 

For the MNIST recognition task, we train our model using RandomForestClassifier (with $5$ folds) from the Python package scikit-learn. For the estimation tasks of the QM-3D dataset, we use the RandomForestRegressor (with $5$ folds) model from the same Python package. For the SKEMPI WT dataset, we use the GradientBoostingRegressor (with $10$ folds to match \cite{Xu2025}) model from the Python package scikit-learn (this corresponds to the PLD-Tree method of \cite{Xu2025}). We use a fixed seed when comparing the results with those obtained using feature vectors from \cite{pmlr-v202-davies23c, Xu2025}. We do not scale our feature vectors in order to present the ideas in and of themselves.

% \subsection{Implementation details}

% Include the practical details that matter for reproducibility: how filtrations were built, what simplicial dimensions were used, how many eigenvalues were computed for the full-spectrum baseline, and how torsion was regularized or stabilized numerically if needed.

% ############################################
% ############################################
\section{Benchmark Results} \label{sec:benchmark}
% ############################################
% ############################################

We compare our feature vectors to the original methods of \cite{pmlr-v202-davies23c, Xu2025}. We expect analogous results  to hold for other models and datasets, which use feature vectors built from (persistent) combinatorial Laplacians.

\subsection{Main predictive performance}

% The main computational results are presented in the following table. 
Table \ref{table:main_results} presents a comparison between our computational results and the computational results of \cite[Table 1]{pmlr-v202-davies23c} and \cite[Table 3, PLD-tree column]{Xu2025}. We use the same quantities for measuring the accuracy as in the original papers.
\begin{table}[h]
\centering
\caption{Main predictive performance: comparison of our compressed topological-spectral representation against baseline full-spectrum or statistical methods from \cite[Table 1]{pmlr-v202-davies23c} and \cite[Table 3, PLD-tree column]{Xu2025}. 
% Standard deviation is provided where appropriate.
Bold values indicate the best performance for each metric. Recall that the SKEMPI WT dataset uses features beyond just the spectral data.}
\begin{tabular}{l p{4cm} c c}
\hline
\textbf{Dataset} & \textbf{Evaluation Metric} & \textbf{Baseline} & \textbf{Our Features} \\
\hline
MNIST & Classification Accuracy & $85.4 \%$ & $\mathbf{86.2}\%$ \\
QM-3D & Mean Absolute Error & $\mathbf{52.2} $ & $53.1$ \\
SKEMPI WT* & Mean Absolute Error & $1.70$ & $\mathbf{1.67}$ \\
SKEMPI WT* & Pearson Correlation & $0.66$ & $\mathbf{0.67}$ \\
\hline
\end{tabular}
\label{table:main_results}
\end{table}

\subsection{Ablation study}\label{subsec:ablation study}

SHAP Feature importance analysis \cite{Lundberg2017AUA} was run on the models using our feature vectors for the MNIST and QM-3D datasets. The analysis showed that the model strongly favored spectral gap and torsion for learning on the QM-3D dataset, and spectral gap and Betti numbers for learning on the MNIST dataset. This shows that all features used contribute to the effectiveness of the feature vector, which can also be confirmed by running a model on only a subset of the features. Table \ref{table:subfeatures} presents the performance of the model on chosen subsets of our features (see Appendix \ref{sec:subset feature}). 

\section{Discussion}
\label{sec:discussion}

\subsection{Stability and Robustness}
The stability of topological descriptors under noise is a central concern in topological data analysis. While persistent homology benefits from well-known stability results, the spectral properties of the persistent Laplacian offer a distinct form of geometric robustness. Analytic torsion, by aggregating the entire non-harmonic spectrum into a single determinant-type quantity, acts as a global summary that is inherently less sensitive to the noise of individual high-frequency eigenvalues. In our benchmarks, this mathematical feature engineering preserves critical signal while bypassing the threshold where increasing raw eigenvalue counts typically leads to diminished model precision.

\subsection{Computational Efficiency}\label{subsec:Computational efficiency}

% A primary motivation for our tripartite representation is the resolution of the spectral packing problem. 
% Extracting a full eigenspectrum for each filtration scale is computationally intensive and leads to relatively long feature vectors. 
 
Our proposed feature vectors tend to be low-dimensional, at the very least halving the length of the existing feature vectors.
This lower dimensionality significantly mitigates the risk of overfitting and avoids the ``breaking point'' where high-frequency noise hinders model precision \cite{pmlr-v202-davies23c}.
Additionally,
This leads to shorter training times, reduced computational overhead, and memory footprint required for large-scale datasets. See the Appendix \ref{sec:exact training} for timing results.

\subsection{Limitations and the Global-Local Tradeoff}\label{sec: limitations}
It is important to acknowledge that analytic torsion is a global invariant. While it effectively summarizes how geometric "twisting" and connectivity evolve across a filtration, it does so by compressing the higher-frequency information into a single scalar. For tasks that depend on highly localized geometric primitives or specific oscillatory modes, the raw non-harmonic spectrum may still contain discriminative information that torsion aggregates away. However, our empirical results suggest that for a wide range of benchmarks in molecular physics and computer vision, this global-local tradeoff strongly favors the compact spectral representation over raw
%\JG{removed adjective}, high-dimensional 
spectral data.

% ############################################
% ############################################
\section{Conclusion}
% ############################################
% ############################################

% Persistent Laplacians enrich topological data analysis by incorporating non-harmonic spectral information, we have  concentrated this information into three main features -- Betti numbers, spectral gaps, and the analytic torsion. Across benchmark datasets, our choice of features performs comparably to the full persistent-Laplacian eigenspectrum, and mostly outperforms the statistical interpretation of the persistent Laplacian spectrum. Hence providing a compact and interpretable spectral summaries of fixed length, removing any guess work required for preparation of feature vectors for learning. This work helps connect classical ideas from spectral topology with modern data-driven pipelines in a way that is both mathematically natural and algorithmically effective.
 
Persistent Laplacians significantly enrich topological data analysis by incorporating non-harmonic spectral information that captures the geometric evolution of filtered complexes. In this work, we have demonstrated that this vast spectral channel can be effectively concentrated into a compact, tripartite feature set: Betti numbers, spectral gaps, and analytic torsion. 

Across benchmark datasets, our proposed representation performs comparably to the full persistent-Laplacian eigenspectrum and consistently outperforms the statistical summaries used in previous studies. By providing a mathematically principled and fixed-length interface, we eliminate the manual optimization and heuristic guesswork typically associated with spectral feature engineering.

This work establishes a transparent framework for characterizing how shape and connectivity evolve across a filtration, effectively bridging the gap between classical spectral geometry and modern topological learning. Future directions will include the investigation of sheaf-based persistent Laplacians \cite{jones2026petlspersistenttopologicallaplacian, wang2026multidimensionalpersistentsheaflaplacians, Wei2025}
 and the stability of these invariants under varying noise regimes, in line with \cite{10.1007/978-981-95-4969-6_30, jung2025persistentlaplaciandiagrams}.

\section*{Data and Software Availability}\label{sec:code}
The source code used for these benchmarks was modified from \url{https://github.com/tomogwen/persistentlaplaciandatascience} (MIT license) and \url{https://github.com/xxjan719/PLNet} (MIT license), corresponding to \cite{pmlr-v202-davies23c, Xu2025} respectively. The modified code will be made available upon request. The raw datasets are sourced from their respective official repositories as cited in Section~\ref{sec:exp}.

\newpage
\begin{ack}
 A.D.L.\ and J.G.\ were partially supported by NSF grants DMS-1902092 and DMS-2200419, and the Simons Foundation collaboration grant on New Structures in Low-dimensional Topology. Computations associated with this project were conducted utilizing the Center for Advanced Research Computing (CARC) at the University of Southern California. We would like to thank H. J. Arai for helpful conversations.
\end{ack}


%%%%%%%%%%%%%%%%%%%%%%%%%%%%%%%%%%%%%%%%%%%%%%%%%%%%%%%%%%%%
\newpage
\appendix

\section{Performance on subsets of features}\label{sec:subset feature}
This appendix is meant as supporting material to Section \ref{subsec:ablation study}. We provide the evaluation of the model on only a subset of our features as well as on the statistical spectral data from \cite{Xu2025}. The results suggest that all of our chosen features contribute to the final feature vector and they have different performance on the different tasks.

\begin{table}[h!]
    \centering
    \caption{Performance of the model on subsets of our features for MNIST and QM-3D datasets, and on the statistical feature vectors. The numbers in the MNIST column represent average percentages of correct predictions, and those in the QM-3D column show the mean absolute error in energy prediction.  } 
\begin{tabular}{lccc}
  \hline
  \textbf{Features}& \textbf{MNIST} & \textbf{QM-3D} \\
  \hline
  Spectral gap & $ 74.2 \%$ &$ 62.1$ \\
  Torsion & $49.0 \%$& $60.2 $   \\
  Betti numbers & $69.0 \%$& $170.8$   \\
  Spectral gap and Betti numbers & $85.2 \%$& $61.4$   \\
  Spectral gap and torsion & $77.0 \%$& $53.7$   \\
  Betti numbers and torsion & $77.1 \%$& $57.2$   \\
 Statistical spectral data as in \cite{Xu2025} & $78.8 \%$&$53.1$  \\
  \hline  
\end{tabular} 
    \label{table:subfeatures}
\end{table}

\section{Exact training times}\label{sec:exact training}

This appendix is meant as supporting material to \ref{subsec:Computational efficiency}. We present the measurements of the total training time required on a single core of a laptop AMD Ryzen 9 8940HX, see Table \ref{table:training time}, demonstrating speedups across the board. The time required for data preparation can be found in \cite{pmlr-v202-davies23c, Xu2025}, and remains largely unchanged by construction.
\begin{table}[h!]
    \centering
\caption{Total training times for models as described in Section \ref{sec:learning protocol} on a single laptop core. Recall that we incorporated data beyond the spectrum to the SKEMPI WT feature vectors, hence offering lesser time improvement.} 
\begin{tabular}{lccc}
  \hline
  \textbf{Dataset}& \textbf{Baseline \cite{pmlr-v202-davies23c, Xu2025}} & \textbf{Our features} \\
  \hline
  QM-3D ($5$-fold) & $143.6 s$  &$\mathbf{24.9s}$ \\
  MNIST ($5$-fold) & $12.6s$& $\mathbf{3.9s}$   \\
  SKEMPI WT* ($10$-fold) & $284.3s$& $\mathbf{242.5s}$   \\
  \hline  
\end{tabular} 
    \label{table:training time}
\end{table}
%%%%%%%%%%%%%%%%%%%%%%%%%%%%%%%%%%%%%%%%%%%%%%%%%%%%%%%%%%%%

\end{document}